\documentclass[11pt]{article}

\usepackage[final]{acl}

\usepackage{times}
\usepackage{latexsym}

\usepackage[T1]{fontenc}

\usepackage[utf8]{inputenc}

\usepackage{microtype}

\usepackage{inconsolata}

\usepackage{graphicx}
\usepackage{booktabs}
\usepackage{pifont}

\usepackage{amssymb}   
\usepackage{amsmath}

\usepackage{makecell}

\title{E$^2$-LLM: Bridging Neural Signals and Interpretable Affective Analysis}

\author{Fei Ma$^{1,*}$, Han Lin$^{2,*}$, Yifan Xie$^{3}$\thanks{Equal Contribution}\thanks{Corresponding Author}, Hongwei Ren$^{4}$,\\
        \textbf{Xiaoyu Shen$^{5}$, Wenbo Ding$^{3}$, Qi Tian$^{1,6}$} \\
        $^1$Guangdong Laboratory of Artificial Intelligence and Digital Economy (SZ) \\
        $^2$Zhejiang University 
        $^3$Tsinghua University
        $^4$Harbin Institute of Technology \\
        $^5$Eastern Institute of Technology 
        $^6$Huawei \\
        \texttt{ivanxie416@gmail.com}}

\begin{document}
\maketitle
\begin{abstract}
Emotion recognition from electroencephalography (EEG) signals remains challenging due to high inter-subject variability, limited labeled data, and the lack of interpretable reasoning in existing approaches. While recent multimodal large language models (MLLMs) have advanced emotion analysis, they have not been adapted to handle the unique spatiotemporal characteristics of neural signals. We present E²-LLM (EEG-to-Emotion Large Language Model), the first MLLM framework for interpretable emotion analysis from EEG. E²-LLM integrates a pretrained EEG encoder with Qwen-based LLMs through learnable projection layers, employing a multi-stage training pipeline that encompasses emotion-discriminative pretraining, cross-modal alignment, and instruction tuning with chain-of-thought reasoning. We design a comprehensive evaluation protocol covering basic emotion prediction, multi-task reasoning, and zero-shot scenario understanding. Experiments on the dataset across seven emotion categories demonstrate that E²-LLM achieves excellent performance on emotion classification, with larger variants showing enhanced reliability and superior zero-shot generalization to complex reasoning scenarios. Our work establishes a new paradigm combining physiological signals with LLM reasoning capabilities, showing that model scaling improves both recognition accuracy and interpretable emotional understanding in affective computing.
\end{abstract}

\section{Introduction}
Multimodal emotion recognition aims to infer human emotional states by integrating heterogeneous signals such as speech, text, facial expressions, and physiological cues~\citep{shou2025multimodal}. By fusing complementary information across modalities, multimodal approaches achieve more robust affective inference than unimodal methods. Deep learning architectures have evolved from CNNs and RNNs to Transformer-based models that align asynchronous multimodal signals using multi-head self-attention~\citep{yao2024advancing,chen2025eeg}. Recent work has explored graph neural networks to model cross-modal dependencies through message passing~\citep{shou2025dynamic}. However, these methods remain limited to fixed emotion taxonomies and struggle with the subjectivity and ambiguity inherent in affective labeling~\citep{wang2024eegpt}, particularly when fine-grained semantic reasoning is required.

Electroencephalography (EEG), a non-invasive neuroimaging technique recording brain electrical activity through scalp electrodes, shows substantial promise in brain-computer interfaces, cognitive assessment, and neurological diagnostics~\citep{babu2025large}. 
However, EEG signal processing faces significant challenges including artifact contamination from eye movements and muscle activity, high inter- and intra-subject variability, and limited large-scale annotated datasets~\citep{huang2023discrepancy}. Meanwhile, Large Language Models (LLMs) such as GPT and BERT have revolutionized natural language processing through transformer architectures that capture sequential dependencies and enable few-shot and zero-shot learning. The self-attention mechanism in LLMs dynamically weights input features and captures long-range dependencies, making it well-suited for modeling EEG's spatiotemporal dynamics. Recent work has explored convergence of LLMs and EEG analysis, with foundation models like LaBraM~\citep{jianglarge} and EEGPT~\citep{wang2024eegpt} using transformer architectures with masked or autoregressive pretraining to learn generalizable EEG representations. For emotion recognition, Multimodal Large Language Models (MLLMs) have achieved remarkable progress by integrating heterogeneous signals through specialized connector modules, enabling end-to-end multimodal emotion reasoning that surpasses traditional models without extensive labeled data~\citep{yang2025omni}.

Despite recent advances, existing EEG-based emotion recognition approaches have several key limitations. First, while LLM-based methods have shown promise in EEG-to-text translation~\citep{mishra2025thought2text} and foundation models~\citep{jiang2024seed}, they lack emotion-specific reasoning capabilities. Recent emotion-focused MLLMs~\citep{yang2025omni} have not been adapted to handle the unique spatiotemporal characteristics of neural signals. 
Second, conventional systems operate as closed-set classifiers that output categorical labels without interpretable rationales, lacking hierarchical cross-modal alignment mechanisms to capture both local cues (micro facial expressions, pitch shifts) and global affective patterns. 
Third, early LLM-based approaches rely solely on textual features, failing to capture the physiological dynamics central to emotional states. 
Although multimodal emotion benchmarks like EMER~\citep{lian2024affectgpt} and MERR~\citep{liu2024hidden} provide reasoning annotations, no prior work has systematically integrated EEG signals with chain-of-thought emotion reasoning through end-to-end instruction tuning to enable interpretable, open-vocabulary emotion analysis from neural data.

To address these limitations, we propose E²-LLM (EEG-to-Emotion Large Language Model), the first multimodal large language model for interpretable emotion analysis from EEG signals. E²-LLM features three key innovations: (1) A hierarchical architecture integrating a pretrained EEG encoder with Qwen3-based LLMs via learnable projection layers that align spatiotemporal EEG dynamics with language embeddings; (2) A multi-stage training pipeline encompassing emotion-discriminative pretraining, cross-modal alignment, and instruction tuning with chain-of-thought reasoning to establish semantic correspondence while enabling interpretable analysis; (3) A comprehensive evaluation protocol covering basic emotion prediction, multi-task reasoning (pairwise comparison, superlative selection, individual matching), and zero-shot scenario reasoning. 
Evaluated on the SEED-VII dataset~\citep{jiang2024seed} across seven emotion categories, E²-LLM obtains excellent performance, with larger variants demonstrating enhanced zero-shot generalization to unseen complex scenarios. Our work establishes a new paradigm combining physiological signals with LLM reasoning, showing that model scaling enhances both recognition accuracy and interpretable emotional understanding in affective computing.

\section{Related Work}
\paragraph{LLM-based EEG Analysis}
The integration of large language models (LLMs) with electroencephalography (EEG) has emerged as a promising research direction. Recent advances span several key areas: foundation models such as LaBraM~\citep{jianglarge} and EEGPT~\citep{wang2024eegpt} adopt transformer-based architectures with masked or autoregressive pretraining to learn generalizable EEG representations across diverse tasks. Instruction-tuned models like Thought2Text~\citep{mishra2025thought2text} further enable open-vocabulary text generation from neural signals. Beyond text, LLMs serve as semantic intermediaries for cross-modal generation, guiding diffusion-based image synthesis~\citep{liu2024hidden,xie2025ccis} and 3D object reconstruction~\citep{deng2025mind2matter} from brain activity. These developments demonstrate the potential of leveraging language model architectures and training paradigms to advance neural signal analysis and brain-computer interface applications.

\paragraph{LLM-based Emotion Analysis}
Large Language Models (LLMs) have shown strong capabilities in emotion analysis through their language understanding and reasoning abilities. 
Early LLM-based approaches focused on textual emotion recognition, where text encoders extract embeddings projected into a unified space for emotion classification~\citep{lei2023instructerc,ma2025generative,ma2025review}. 
These methods achieved promising zero-shot and few-shot performance using prompting strategies such as instruction following, in-context learning, and chain-of-thought reasoning~\citep{wei2022chain}. 
However, their reliance on text alone limits their ability to capture critical affective cues from visual and acoustic modalities, such as facial micro-expressions~\citep{xie2025pointtalk,lin2024video,xie2025audio} and prosodic variations~\citep{feng2025unisync,feng2025stftcodec}. 
This has motivated the development of multimodal large language models (MLLMs) that integrate heterogeneous signals through specialized connector modules for end-to-end multimodal emotion recognition and reasoning~\citep{yang2025omni}.

\paragraph{Multimodal Large Language Models}
The emergence of Multimodal Large Language Models has marked a pivotal shift in AI, introducing capabilities that span both language and vision. The field's evolution began with leveraging LLMs as coordination systems for task-specific applications~\citep{shen2023hugginggpt,yang2023gpt4tools,xie2025universal}, then progressed toward lightweight adaptation techniques~\citep{hu2022lora} and instruction-driven alignment strategies~\citep{liu2024improved} that connect visual and textual semantics.
In our work, we develop the first large language model capable of bridging EEG signals and emotion analysis, addressing the challenge of automated emotion recognition from neural data.

\section{Method}

\begin{figure*}[t]
    \centering
    \includegraphics[width=\textwidth]{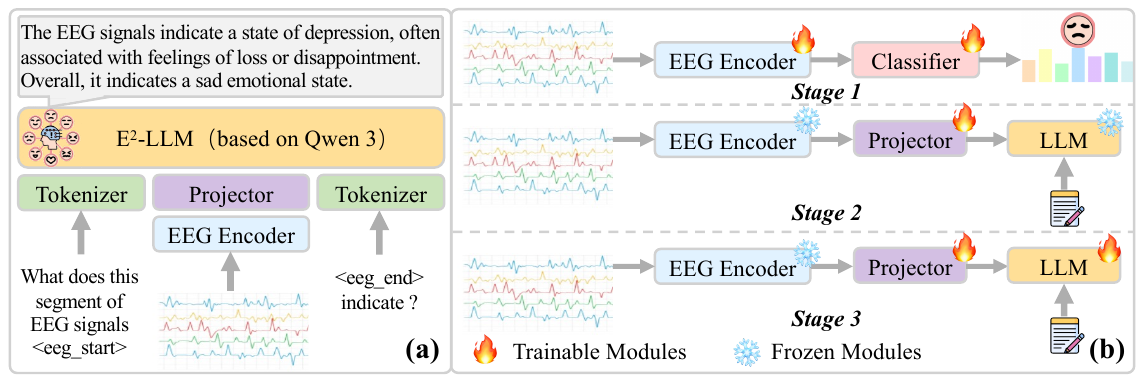}
    \caption{
    Overview of the E²-LLM framework and training pipeline.
    (a) E$^2$-LLM framework: EEG signals are processed by an EEG Encoder, whose representations are mapped to the embedding space of a Qwen3-based LLM via a Projector. Special tokens $\texttt{<eeg\_start>}$ and $\texttt{<eeg\_end>}$ demarcate EEG segments within the input sequence, enabling the LLM to generate interpretive emotional analysis.
    (b) Multi-Stage Training Strategy: Stage 1 trains the EEG Encoder with a classification objective for emotion recognition. Stage 2 freezes the encoder and trains the Projector to align EEG representations with LLM embeddings. Stage 3 jointly fine-tunes the Projector and LLM for generating natural language emotional analysis reports.}
    \label{fig:main}
\end{figure*}

\begin{table}[t]
\centering
\small
\resizebox{\columnwidth}{!}{
\begin{tabular}{p{2.2cm}p{5.8cm}cc}
\toprule
\textbf{Task} & \textbf{Example} & \textbf{Train} & \textbf{Eval} \\
\midrule
Individual Emotion Description (IED) & 
Describe the emotion of \texttt{<EEG>}. & 
\checkmark & \checkmark \\
\midrule
Emotion Pairwise Comparison (EPC) & 
Is the individual with \texttt{<EEG 1>} happier than the individual with \texttt{<EEG 2>}? & 
\checkmark & \checkmark \\
\midrule
Emotion Superlative Selection (ESS) & 
Given three EEG segments: a) \texttt{<EEG 1>}, b) \texttt{<EEG 2>}, c) \texttt{<EEG 3>}, select the saddest. & 
\checkmark & \checkmark \\
\midrule
Emotion Individual Matching (EIM) & 
Match three EEG segments to: 1) someone waiting for a bus, 2) a parent holding their newborn, 3) a person startled by loud noise. & 
\checkmark & \checkmark \\
\midrule
Emotion Scenario Reasoning (ESR) & 
\textit{T1:} Describe three EEG segments. \textit{T2:} [Scenario description] Select the most appropriate segment with justification. & 
\ding{55} & \checkmark \\
\bottomrule
\end{tabular}
}
\caption{Overview of five task types for EEG-based emotion understanding.}
\label{tab:tasks}
\end{table}

\subsection{Data Construction}


We construct our training and evaluation data based on the SEED-VII dataset~\citep{jiang2024seed}, a multimodal emotion recognition benchmark containing EEG signals from 20 subjects across seven emotion categories: happiness, surprise, neutrality, disgust, fear, sadness, and anger. The original dataset provides 62-channel EEG recordings sampled at 1000 Hz.

We first apply a 0.1–70 Hz bandpass filter and a 50 Hz notch filter to these raw signals. In alignment with EEGPT~\citep{wang2024eegpt}, we utilize 58 designated electrodes. The EEG signals are resampled to 256 Hz and partitioned into 10-second segments. Subsequently, they are scaled to mV, re-referenced using a global average reference, and extracted into 4-second windows using a random cropping strategy. Following this preprocessing pipeline, we obtain a total of 24,444 training segments and 2,761 test segments organized by emotion category.

To enable the LLM to perform interpretive emotional analysis from EEG signals, we design a diverse set of question-answer templates spanning five task types with increasing complexity, as summarized in Table~\ref{tab:tasks}.

\textbf{Individual Emotion Description (IED)}: Given a single EEG segment, the model must describe the underlying emotional state (e.g., "Describe the emotion of $\texttt{<EEG>}$."). The target response provides both an unstructured interpretation (e.g., "The EEG signals indicate a state of depression, often associated with feelings of loss or disappointment") and an explicit emotion label.

\textbf{Emotion Pairwise Comparison (EPC)}: The model receives two EEG segments and must determine comparative emotional intensity (e.g., "Is the individual with EEG signals $\texttt{<EEG 1>}$ happier than the individual with $\texttt{<EEG 2>}$?"). This requires the model to first describe both emotional states before drawing a comparative conclusion.

\textbf{Emotion Superlative Selection (ESS)}: Given three EEG segments, the model identifies which exhibits the strongest or weakest manifestation of a target emotion (e.g., "Given three segments of EEG signals: a) $\texttt{<EEG 1>}$, b) $\texttt{<EEG 2>}$, c) $\texttt{<EEG 3>}$, select the saddest."), requiring multi-sample reasoning and ranking.

\textbf{Emotion Individual Matching (EIM)}: The model matches three EEG segments to scenario descriptions of individuals experiencing different emotions (e.g., matching segments to "someone waiting for a bus on an ordinary afternoon," "a new parent holding their sleeping newborn," and "a person startled by an abrupt, loud noise"), evaluating the model's ability to ground physiological signals in real-world emotional contexts.

\textbf{Emotion Scenario Reasoning (ESR)}: This two-turn task first requires the model to describe three EEG segments, then presents a nuanced scenario (e.g., "At a family gathering, your aunt sets the table with one fewer plate than usual. No one comments, but your cousin quietly takes the extra chair away...") and asks the model to select the most appropriate segment with justification. This task evaluates complex reasoning by requiring implicit emotion inference from contextual cues.

We generate 10,000 training samples using the first four task types (IED, EPC, ESS, EIM) derived from the processed training segments, ensuring balanced representation across emotion categories. For evaluation, we utilize the 2,761 held-out test segments to construct 2,761 IED samples, 500 multi-task reasoning (EPC, ESS, and EIM) samples, and 167 ESR samples. These are held out from training to assess the model's generalization to complex emotional reasoning scenarios. All question-answer pairs follow a chain-of-thought format, first describing individual emotional states before providing final conclusions, which encourages interpretable reasoning during inference.

\subsection{E$^2$-LLM}

\subsubsection{Overview}
Figure~\ref{fig:main} illustrates the overall architecture of E²-LLM, which comprises three core components: an EEG Encoder based on EEGPT~\citep{wang2024eegpt}, a Projector, and a Qwen3~\citep{yang2025qwen3}-based Large Language Model (LLM). The framework processes EEG signals to generate interpretive emotional analysis through a unified multimodal architecture.

\paragraph{EEG Signal Processing} Given an input EEG signal $\mathbf{X} \in \mathbb{R}^{M \times T}$ with $M$ channels and $T$ time points, we first segment it into non-overlapping patches $\mathbf{p}_{i,j}$ along both spatial and temporal dimensions:
\begin{equation}
\begin{aligned}
& \mathbf{p}_{i,j} = \mathbf{X}_{i,(j-1)d:jd}, \\
& \quad i \in \{1,\ldots,M\}, 
j \in \{1,\ldots,N\}
\end{aligned}
\end{equation}

where $d$ denotes the temporal patch length and $N = T/d$ represents the number of temporal patches. Each patch is then embedded through a local spatio-temporal embedding layer:
\begin{equation}
\mathbf{e}_{i,j} = \mathbf{W}_p^\top \mathbf{p}_{i,j} + \mathbf{b}_p + \boldsymbol{\varsigma}_i
\end{equation}
where $\mathbf{W}_p \in \mathbb{R}^{d \times d_e}$ and $\mathbf{b}_p \in \mathbb{R}^{d_e}$ are learnable parameters, and $\boldsymbol{\varsigma}_i$ denotes the channel-specific embedding retrieved from a learnable codebook.

\paragraph{Hierarchical Encoding} The EEG Encoder adopts a hierarchical transformer architecture that processes spatial and temporal information separately. For each time step $j$, the encoder aggregates spatial information across channels:
\begin{equation}
\mathbf{h}_j = \text{ENC}\left({\mathbf{e}_{i,j}}_{i=1}^{M}\right)
\end{equation}
This design reduces computational complexity from $\mathcal{O}((M \times N)^2)$ to $\mathcal{O}(M^2 \times N)$ while enhancing flexibility for varying electrode configurations.

\paragraph{Multimodal Alignment} The Projector maps EEG representations into the embedding space of the language model. We employ a two-layer MLP with GELU activation:
\begin{equation}
\mathbf{z} = \mathbf{W}_2 \cdot \text{GELU}(\mathbf{W}_1 \cdot \mathbf{h} + \mathbf{b}_1) + \mathbf{b}_2
\end{equation}
where $\mathbf{h}$ denotes the encoder output and $\mathbf{z} \in \mathbb{R}^{d{\text{LLM}}}$ represents the projected embedding aligned with the LLM's token space.

\paragraph{Input Formulation} We introduce special tokens $\texttt{<eeg\_start>}$ and $\texttt{<eeg\_end>}$ to demarcate EEG segments within the input sequence. The final input to LLM is constructed as:
\begin{equation}
\begin{aligned}
\mathbf{S} = [& \mathbf{w}_1,\dots,\mathbf{w}_k,\text{\texttt{<eeg\_start>}},\mathbf{z}_1,\dots,\mathbf{z}_N, \\
& \text{\texttt{<eeg\_end>}},\mathbf{w}_{k+1},\dots]
\end{aligned}
\end{equation}
where ${\mathbf{w}_i}$ are text token embeddings and ${\mathbf{z}_j}$ are projected EEG embeddings, enabling the LLM to generate natural language emotional analysis reports conditioned on both textual instructions and EEG signals.

\subsubsection{Multi-stage Training Strategy}
As illustrated in Figure~\ref{fig:main}(b), we adopt a three-stage training pipeline to progressively bridge the modality gap between EEG signals and natural language. This curriculum-based approach ensures stable optimization and effective cross-modal alignment.

\paragraph{Stage 1: EEG Encoder Training} In the first stage, we fine-tune the pretrained EEG Encoder with a classification objective to learn discriminative emotion representations. Given the encoded features $\mathbf{h} = \text{ENC}(\mathbf{X})$, a classification head predicts the emotion category:
\begin{equation}
\hat{\mathbf{y}} = \text{Softmax}(\mathbf{W}_c \cdot \mathbf{h} + \mathbf{b}_c)
\end{equation}
where $\mathbf{W}_c \in \mathbb{R}^{C \times d_e}$ and $\mathbf{b}_c \in \mathbb{R}^C$ are learnable parameters, and $C$ denotes the number of emotion categories. The encoder is optimized using cross-entropy loss:
\begin{equation}
\mathcal{L}_{\text{cls}} = -\sum_{c=1}^C y_c \log(\hat{y}_c)
\end{equation}
where $y_c$ represents the ground-truth label. During this stage, both the Projector and LLM modules remain frozen, allowing the encoder to focus on extracting emotion-relevant features from raw EEG signals.

\paragraph{Stage 2: Cross-modal Alignment} In the second stage, we freeze the EEG Encoder and train the Projector to align EEG representations with the LLM's embedding space. The model is trained using the autoregressive language modeling objective:
\begin{equation}
\mathcal{L}_{\text{LM}} = -\sum_{l=1}^L \log P_{\theta}(w_l|\mathbf{S}_{<l})
\end{equation}
where $L$ is the length of the target response, $w_l$ denotes the $l$-th token, and $\mathbf{S}_{<l}$ represents all preceding tokens including both textual instructions and projected EEG embeddings. This stage establishes a semantic bridge between the two modalities while preserving the learned EEG representations.

\paragraph{Stage 3: End-to-end Fine-tuning} In the final stage, we jointly fine-tune the Projector and LLM while keeping the EEG Encoder frozen. The model continues to be optimized with the same language modeling objective $\mathcal{L}_{\text{LM}}$, but now updates both the Projector and LLM parameters. To enhance parameter efficiency, we employ Low-Rank Adaptation (LoRA)~\citep{hu2022lora,zhu2024minigpt} for the LLM, which significantly reduces the number of trainable parameters while maintaining model expressiveness. This stage enables the model to generate fluent and accurate natural language emotional analysis reports conditioned on EEG inputs.

\section{Experiments}

\begin{figure*}[t]
    \centering
    \includegraphics[width=\textwidth]{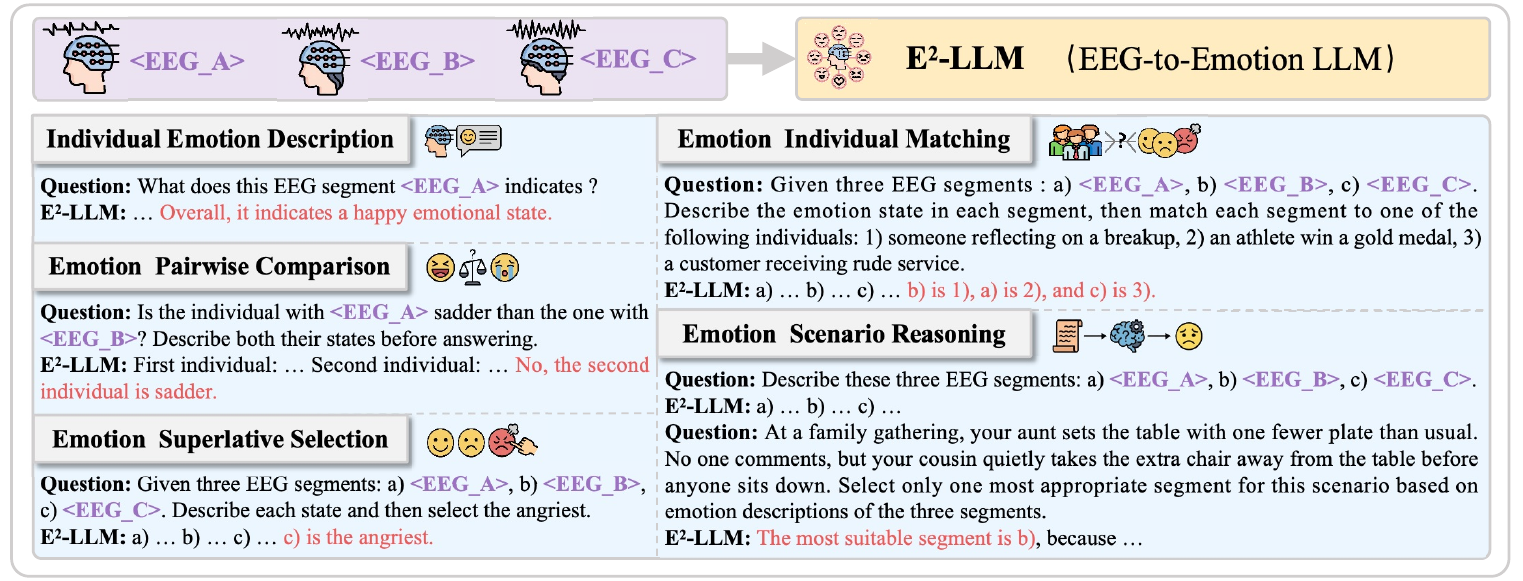}
    \caption{
    Illustrative examples from the proposed E$^2$-LLM across five distinct tasks. The qualitative results demonstrate that E$^2$-LLM can generate reasonable outputs. 
    }
    \label{fig:example}
\end{figure*}

\begin{table*}[t]
\centering
\small
\begin{tabular*}{\textwidth}{@{\hspace{\tabcolsep}\extracolsep{\fill}}l|ccccc}
\toprule
 & Random & NeuroLM & \textbf{E$^2$-LLM$_\text{4B}$} & \textbf{E$^2$-LLM$_\text{8B}$} & \textbf{E$^2$-LLM$_\text{14B}$} \\
\midrule
"Happy" & 14.29 & 11.71 & 33.02 & 46.42 & \textbf{51.40} \\
"Surprise" & 14.29 & 35.62 & 71.82 & 65.34 & \textbf{76.20} \\
"Neutral" & 14.29 & 18.24 & 49.01 & \textbf{80.24} & 70.75 \\
"Disgust" & 14.29 & 14.30 & 8.24 & \textbf{45.88} & 20.88 \\
"Fear" & 14.29 & 40.90 & \textbf{77.48} & 75.00 & 76.35 \\
"Sad" & 14.29 & 35.02 & \textbf{75.37} & 46.56 & 53.65 \\
"Anger" & 14.29 & 21.97 & 67.46 & 72.68 & \textbf{78.15} \\
\midrule
\textbf{Balanced Accuracy} & 14.29 & 25.39 & 54.63 & \textbf{61.73} & 61.05 \\
\textbf{Cohen's Kappa} & 0.00 & 13.65 & 49.69 & 54.72 & \textbf{55.14} \\
\textbf{Weighted F1-score} & 14.29 & 25.83 & 54.56 & \textbf{60.91} & 60.16 \\
\bottomrule
\end{tabular*}
\caption{Performance (\%) comparison on the Individual Emotion Description (IED) task. Results demonstrate the effectiveness of E$^2$-LLM variants against baselines across seven emotion categories. Cohen’s kappa
measures prediction reliability by accounting for chance agreement. Best results are \textbf{bolded}.}
\label{tab:ied}
\end{table*}

\subsection{Experimental Setup}



Our architecture incorporates the 10M-parameter EEG encoder from EEGPT~\citep{wang2024eegpt} and adopt Qwen3~\citep{yang2025qwen3} as the backbone language model. We train and evaluate three variants of the proposed framework---E$^2$-LLM$_\text{4B}$, E$^2$-LLM$_\text{8B}$, and E$^2$-LLM$_\text{14B}$---employing Qwen3-4B, Qwen3-8B, and Qwen3-14B respectively.

\paragraph{Training Details} Encoder fine-tuning was conducted on the 24,444 training segments for 50 epochs using the AdamW optimizer ~\citep{loshchilov2017decoupled} with a weight decay of 0.01, a learning rate of $10^{-4}$, batch size of 64, following a one-cycle learning rate schedule. For cross-modal alignment, the Projector module was trained on 10,000 training question-answer samples using Adam~\citep{kingma2014adam} with no weight decay, a learning rate of $2 \times 10^{-4}$, and a batch size of 16 under a cosine annealing schedule. 

For end-to-end fine-tuning, both the Projector and the LoRA~\citep{hu2022lora} parameters of the LLM were optimized using 3,000 training samples. We maintained a batch size of 16 and a learning rate of $2 \times 10^{-4}$ for both components, utilizing Adam under a cosine annealing schedule without weight decay. The LoRA configuration involved a scaling factor of 256, a rank of 128, and a dropout rate of 0.05, targeting the Query Projection Layer ($W_Q$) and the Key Projection Layer ($W_K$) of the language model.

Regarding computational costs, encoder fine-tuning took less than 7 hours and required less than 34 GB of GPU VRAM. The combined cross-modal alignment and end-to-end fine-tuning stages required 2.3, 3.2, and 5.4 hours for the 4B, 8B, and 14B variants, respectively. Experiments for the 4B model were conducted on a single NVIDIA RTX 6000 Ada Generation GPU, while the 8B and 14B models were trained using two such GPUs.

\paragraph{Evaluation Protocol}
To rigorously evaluate the performance of E$^2$-LLM, we utilize the test QA pairs derived from the held-out test segments. The evaluation framework spans three primary dimensions: 
1) For the basic emotion prediction task (IED), we report balanced accuracy, Cohen's kappa, and weighted F1-score across the entire test set to provide a robust measure of classification performance. 
2) We measure the accuracy on multi-task reasoning questions (EPC, ESS, and EIM) to assess the model's multi-faceted cognitive reasoning capabilities. 
3) The model's generalization ability is quantified via accuracy over a specialized scenario-based reasoning task (ESR) that remains unseen during training.

Visual illustrations of the five tasks mentioned above and the corresponding model responses are provided in Figure~\ref{fig:example}.

\subsection{Results}

\paragraph{Emotion Prediction}

Table~\ref{tab:ied} summarizes the performance of the E$^2$-LLM variants and baseline methods on the individual emotion description (IED) task. Crucially, we benchmark against NeuroLM~\citep{jiang2024neurolm}, a recent paradigm that similarly couples EEG encoder and LLM. Following its protocol, we fine-tuned NeuroLM in different parameter scales on SEED-VII~\citep{jiang2024seed} and report the best-performing version. However, it achieves a balanced accuracy of only 25.39\%. Further analysis reveals that NeuroLM suffers from overfitting to the rigid instruction formats used during classification training, rendering it unable to generalize to varied instruction forms.

In contrast, E$^2$-LLM demonstrates superior performance across all metrics, with the benefits of scaling the LLM backbone becoming immediately apparent. Specifically, the 8B model offers a substantial improvement over the 4B variant, attaining the highest balanced accuracy (61.73\%) and weighted F1-score (60.91\%). Notably, while the 14B variant shows a marginal decrease in balanced accuracy, it secures the superior Cohen's kappa value (55.14\%). Given that Cohen's kappa is a robust measure of inter-rater agreement that explicitly corrects for chance, this result underscores that the 14B model provides the most consistent and reliable prediction quality. These observations suggest that scaling the LLM backbone enhances the intrinsic reliability of EEG-based emotion decoding, establishing larger variants as more robust foundations for classification tasks.

\paragraph{Multi-task Reasoning} 

\begin{table}[t]
\centering
\small
\begin{tabular*}{\columnwidth}{@{\hspace{\tabcolsep}\extracolsep{\fill}}l|cccc|c}
\toprule
 & \textbf{EPC} & \textbf{ESS} & \textbf{EIM} & \textbf{ESR*} & \textbf{Avg.} \\
\midrule
Random & 33.33 & 33.33 & 16.67 & 33.33 & 29.17 \\
\textbf{E$^2$-LLM$_\text{4B}$} & 67.48 & 72.94 & \textbf{77.84} & 34.13 & 63.10 \\
\textbf{E$^2$-LLM$_\text{8B}$} & 68.71 & 76.47 & 71.86 & 41.92 & 64.74 \\
\textbf{E$^2$-LLM$_\text{14B}$} & \textbf{72.39} & \textbf{79.41} & 73.65 & \textbf{53.89} & \textbf{69.84} \\
\bottomrule
\end{tabular*}
\caption{Accuracy (\%) results on the EPC, ESS, EIM, and ESR tasks. The ESR task, marked with *, evaluates zero-shot generalization to unseen complex scenarios. Best results are \textbf{bolded}.}
\label{tab:multitask}
\end{table}

Table~\ref{tab:multitask} presents the evaluation results for multi-task reasoning capabilities, encompassing Emotion Pairwise Comparison (EPC), Emotion Superlative Selection (ESS), and Emotion Individual Matching (EIM). All E$^2$-LLM models drastically surpass the corresponding random baselines across these complex cognitive tasks, decisively validating the framework's ability to extract and utilize EEG information for sophisticated relational and contextual reasoning. In general, model scaling correlates positively with enhanced reasoning capability, with E$^2$-LLM$_\text{14B}$ achieving the highest overall average accuracy. This outcome supports the hypothesis that increased LLM capacity provides a superior foundation for handling intricate, diverse cognitive demands. 

However, a deviation is observed: E$^2$-LLM$_\text{4B}$ achieves the peak performance on the EIM task (77.84\%), surpassing its larger 8B and 14B counterparts. This result suggests that a smaller model size may occasionally find a efficient, task-specific optimal mapping for tasks requiring focused semantic matching, which may be slightly diluted by the broader reasoning capacities optimized in the larger models.

\paragraph{Scenario Reasoning}

The Emotion Scenario Reasoning (ESR) task serves as a critical test of generalization, as its complex two-turn format and contextual inference requirements were unseen during training. As shown in Table~\ref{tab:multitask}, this task reveals a distinct phenomenon linked to model scale. The 4B model cannot generalize well, yielding an accuracy of 34.13\%, which is just slightly higher than the random baseline. Qualitative inspection indicates that the 4B model sometimes struggles to adhere to the novel instruction format, often failing to generate valid responses---a limitation typical of smaller models that overfit to fixed training templates, which is even much more severe in NeuroLM~\citep{jiang2024neurolm}. Conversely, increasing the model size to 8B and 14B unlocks significant zero-shot generalization capabilities, with performance jumping to 41.92\% and 53.89\% respectively. This finding suggests that while smaller models can master specific EEG-to-text mappings, a larger language backbone is indispensable for flexibly applying these representations to unseen, high-level reasoning scenarios.


\subsection{Ablation Studies}


\begin{figure}[t]
    \centering
    \includegraphics[width=\columnwidth]{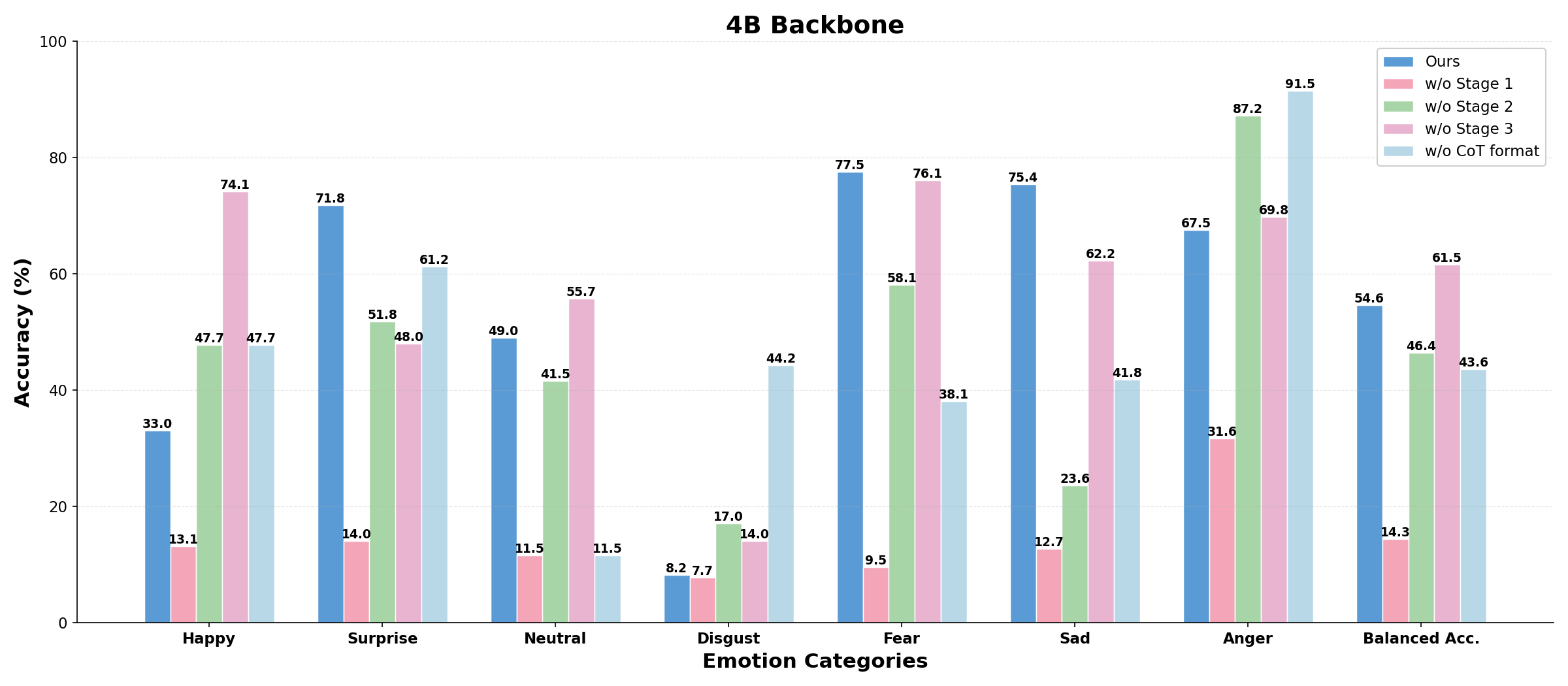}\\[0.5em]
    \includegraphics[width=\columnwidth]{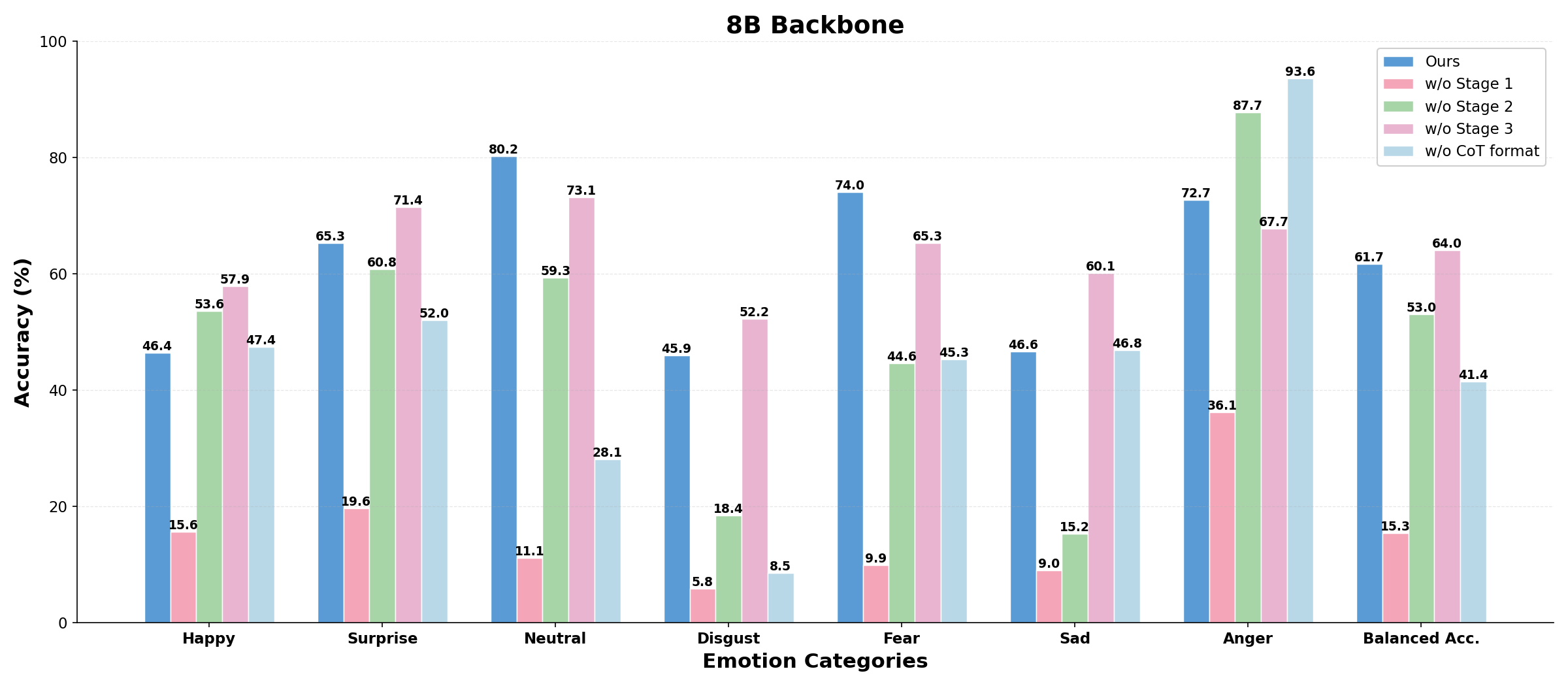}\\[0.5em]
    \includegraphics[width=\columnwidth]{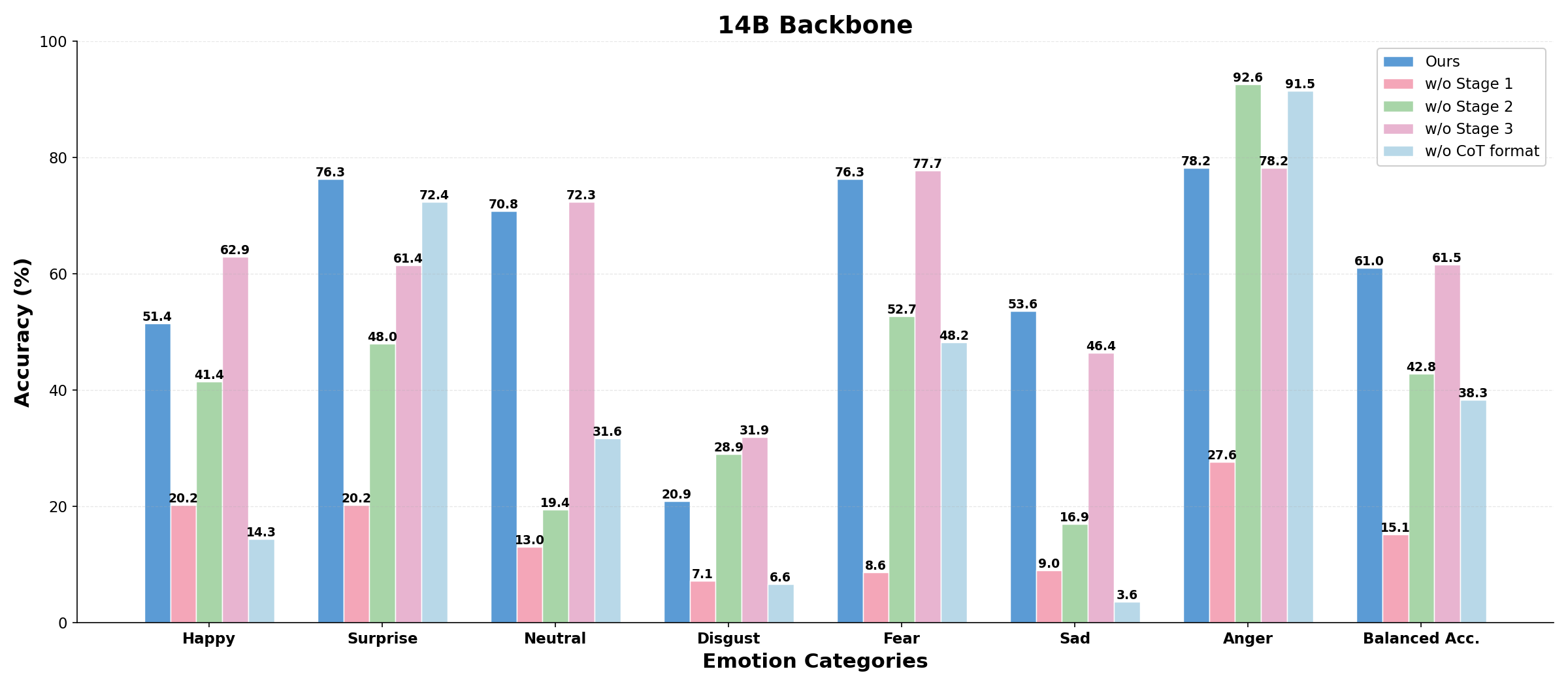}
    \caption{Ablation analysis on the IED task across different model scales.}
    \label{fig:cls}
\end{figure}

\begin{table}[t]
\centering
\small
\begin{tabular*}{\columnwidth}{@{\hspace{\tabcolsep}\extracolsep{\fill}}l|cccc|c}
\toprule
 & \textbf{EPC} & \textbf{ESS} & \textbf{EIM} & \textbf{ESR} & \textbf{Avg.} \\
\midrule
\textbf{E$^2$-LLM$_\text{4B}$} & \textbf{67.5} & \textbf{72.9} & \textbf{77.8} & 34.1 & \textbf{63.1} \\
w/o Stage 1 & 32.5 & 28.2 & 14.4 & 7.2 & 20.6 \\
w/o Stage 2 & 66.3 & 72.9 & 66.5 & 37.1 & 60.7 \\
w/o Stage 3 & 65.6 & 70.0 & 68.9 & 15.0 & 54.9 \\
w/o CoT format & 55.8 & 65.3 & 62.9 & \textbf{38.3} & 55.6 \\
\midrule
\textbf{E$^2$-LLM$_\text{8B}$} & 68.7 & 76.5 & \textbf{71.9} & \textbf{41.9} & \textbf{64.7} \\
w/o Stage 1 & 42.3 & 28.8 & 17.4 & 28.7 & 29.3 \\
w/o Stage 2 & 68.7 & \textbf{77.1} & 69.5 & 37.7 & 63.3 \\
w/o Stage 3 & \textbf{74.2} & 73.5 & 59.9 & 25.7 & 58.3 \\
w/o CoT format & 53.4 & 31.2 & 21.0 & 40.1 & 36.4 \\
\midrule
\textbf{E$^2$-LLM$_\text{14B}$} & \textbf{72.4} & \textbf{79.4} & 73.7 & \textbf{53.9} & \textbf{69.8} \\
w/o Stage 1 & 39.9 & 30.6 & 21.0 & 31.7 & 30.8 \\
w/o Stage 2 & 67.5 & 75.9 & 71.3 & 40.1 & 63.7 \\
w/o Stage 3 & 67.5 & 65.9 & 64.7 & 38.9 & 59.3 \\
w/o CoT format & 58.9 & 69.4 & \textbf{76.6} & 50.9 & 64.0 \\
\bottomrule
\end{tabular*}
\caption{Ablation analysis on the EPC, ESS, EIM, and ESR tasks across different model scales. 
We \textbf{bold} the best for each backbone.}
\label{tab:ablation}
\end{table}

To validate the contribution of each training stage and the chain-of-thought question-answer format, we conduct ablation studies on the five tasks (IED, EPC, ESS, EIM, and ESR), with results summarized in Figure~\ref{fig:cls} and Table~\ref{tab:ablation}. 

We analyze four variants: 1) "w/o Stage 1": Using the pretrained EEGPT encoder~\citep{wang2024eegpt} directly instead of the training it with a classification objective for emotion recognition; 2) "w/o Stage 2": Skipping the separate alignment training for the Projector module and proceeding directly to end-to-end fine-tuning; 3) "w/o Stage 3": Using the model after the alignment stage without end-to-end instruction tuning; and 4) "w/o CoT format": Training and evaluating the model without the prompts and reference answers that guide the LLM to analyze emotion states before providing conclusion.

\paragraph{Impact of Multi-stage Training}
The results demonstrate that the proposed multi-stage training strategy is critical for effective emotion recognition and reasoning.
First, removing the emotion-discriminative training for the EEG encoder ("w/o Stage 1") leads to a catastrophic performance drop across all tasks and backbone sizes, with balanced accuracy falling to near-chance levels. This indicates that the LLM cannot compensate for a "blind" encoder, which means the EEG encoder must learn robust emotion-specific representations before being aligned with the LLM.
Second, the cross-modal alignment is equally vital---the "w/o Stage 2" variant suffers a significant performance degradation (e.g., dropping from 61.0\% to 42.8\% for the 14B model on the IED task). This confirms that the Projector requires a dedicated alignment phase to map physiological signals into the semantic space of the LLM.

An interesting observation from the IED task is that the model without instruction tuning ("w/o Stage 3") achieves classification accuracy comparable to, or marginally higher than, the full model (e.g., 61.5\% vs. 61.0\% for the 14B backbone). This suggests that for simple classification tasks (IED), the alignment learned in Stage 2 is sufficient. However, as shown in Table~\ref{tab:ablation}, the "w/o Stage 3" model fails significantly on more complex reasoning tasks (EPC, ESS, EIM, and ESR), lacking the instruction-following capabilities required to compare emotions or associate emotions with context. Stage 3 is therefore essential for generalizing from simple recognition to interpretable analysis, even if it introduces a slight trade-off in pure classification metrics due to the increased complexity of the generation objective.

\paragraph{Efficacy of Chain-of-Thought} The removal of the chain-of-thought format in both training and evaluation question-answer samples ("w/o CoT format") results in a sharp decline in overall accuracy across all scales (e.g., an average drop of almost 10\% for the 14B model across five tasks). This validates our hypothesis that decomposing the task---forcing the model to analyze and describe the emotion state of EEG before predicting the label or selecting the answer---serves as a crucial regularizer, enabling the LLM to ground its predictions in the observed signal dynamics rather than hallucinating answers.

\section{Conclusion}
We present E²-LLM, the first multimodal large language model framework that bridges EEG signals with interpretable emotion analysis. By integrating a pretrained EEG encoder with Qwen-based LLMs through learnable projections and a multi-stage training pipeline, E²-LLM achieves strong emotion recognition performance while providing human-interpretable explanations of affective states. Our evaluation demonstrates that model scaling enhances both classification accuracy and zero-shot generalization to complex reasoning scenarios. This work establishes a new paradigm combining physiological signals with LLM reasoning capabilities, showing that neural dynamics can be effectively translated into semantic emotional understanding. Future directions include incorporating additional modalities, improving cross-dataset generalization ability, and addressing the inherent subjectivity in emotion labeling for real-world deployment.





\section*{Limitation}
Our study has several limitations that should be acknowledged. First, experiments are conducted solely on the SEED-VII dataset with only 20 subjects, which may limit generalizability across diverse populations and neurophysiological variations. Second, the larger model variants (8B and 14B) require substantial computational resources, potentially limiting accessibility for researchers with constrained budgets. Third, we observe inconsistent scaling behaviors across tasks---notably, the smaller 4B model outperforms larger variants on the Emotion Individual Matching task, suggesting non-monotonic relationships between model size and performance. Fourth, even our best-performing model achieves only 53.89\% accuracy on zero-shot scenario reasoning, indicating significant room for improvement in generalizing to unseen complex emotional contexts. Finally, while E²-LLM generates natural language explanations, we lack systematic human evaluation of the semantic quality and clinical relevance of these interpretations, which remains an important direction for establishing rigorous assessment metrics.

\section*{Ethical Concern}
This research involves analysis of EEG signals for emotion recognition, which raises important ethical considerations regarding privacy and potential misuse. EEG data contains sensitive neurophysiological information that could reveal private mental states beyond intended emotion categories. While our experiments use the publicly available SEED-VII dataset with appropriate informed consent, we acknowledge that automated emotion recognition systems could be misused for unauthorized surveillance, workplace monitoring, or discriminatory decision-making without explicit user consent. Furthermore, the model's interpretive capabilities could potentially generate seemingly authoritative but inaccurate psychological assessments. We strongly advocate that deployment of EEG-based emotion recognition must be governed by strict ethical guidelines, including explicit informed consent, transparent data usage disclosure, robust privacy protections, and safeguards against coercive applications. Future work should prioritize developing technical safeguards and regulatory frameworks to ensure neural signal analysis respects human dignity and individual rights.


\bibliography{custom}

@article{jiang2024seed,
  title={Seed-vii: A multimodal dataset of six basic emotions with continuous labels for emotion recognition},
  author={Jiang, Wei-Bang and Liu, Xuan-Hao and Zheng, Wei-Long and Lu, Bao-Liang},
  journal={IEEE Transactions on Affective Computing},
  year={2024},
  publisher={IEEE}
}

@inproceedings{zhu2024minigpt,
  title={MiniGPT-4: Enhancing Vision-Language Understanding with Advanced Large Language Models},
  author={Zhu, Deyao and Chen, Jun and Shen, Xiaoqian and Li, Xiang and Elhoseiny, Mohamed},
  booktitle={ICLR},
  year={2024}
}

@article{hu2022lora,
  title={Lora: Low-rank adaptation of large language models.},
  author={Hu, Edward J and Shen, Yelong and Wallis, Phillip and Allen-Zhu, Zeyuan and Li, Yuanzhi and Wang, Shean and Wang, Lu and Chen, Weizhu and others},
  journal={ICLR},
  volume={1},
  number={2},
  pages={3},
  year={2022}
}

@article{yang2025omni,
  title={Omni-emotion: Extending video mllm with detailed face and audio modeling for multimodal emotion analysis},
  author={Yang, Qize and Bai, Detao and Peng, Yi-Xing and Wei, Xihan},
  journal={arXiv preprint arXiv:2501.09502},
  year={2025}
}

@article{feng2025unisync,
  title={UniSync: A Unified Framework for Audio-Visual Synchronization},
  author={Feng, Tao and Xie, Yifan and Guan, Xun and Song, Jiyuan and Liu, Zhou and Ma, Fei and Yu, Fei},
  journal={arXiv preprint arXiv:2503.16357},
  year={2025}
}

@inproceedings{lin2024video,
  title={Video-llava: Learning united visual representation by alignment before projection},
  author={Lin, Bin and Ye, Yang and Zhu, Bin and Cui, Jiaxi and Ning, Munan and Jin, Peng and Yuan, Li},
  booktitle={Proceedings of the 2024 conference on empirical methods in natural language processing},
  pages={5971--5984},
  year={2024}
}

@article{wei2022chain,
  title={Chain-of-thought prompting elicits reasoning in large language models},
  author={Wei, Jason and Wang, Xuezhi and Schuurmans, Dale and Bosma, Maarten and Xia, Fei and Chi, Ed and Le, Quoc V and Zhou, Denny and others},
  journal={Advances in neural information processing systems},
  volume={35},
  pages={24824--24837},
  year={2022}
}

@article{ma2025review,
  title={A review of human emotion synthesis based on generative technology},
  author={Ma, Fei and Xie, Yifan and Li, Yukan and He, Ying and Zhang, Yi and Ren, Hongwei and Liu, Zhou and Yao, Wei and Ren, Fuji and Yu, Fei Richard and others},
  journal={IEEE Transactions on Affective Computing},
  year={2025},
  publisher={IEEE}
}

@article{lei2023instructerc,
  title={Instructerc: Reforming emotion recognition in conversation with multi-task retrieval-augmented large language models},
  author={Lei, Shanglin and Dong, Guanting and Wang, Xiaoping and Wang, Keheng and Qiao, Runqi and Wang, Sirui},
  journal={arXiv preprint arXiv:2309.11911},
  year={2023}
}

@inproceedings{xie2025ccis,
  title={Ccis-diff: A generative model with stable diffusion prior for controlled colonoscopy image synthesis},
  author={Xie, Yifan and Wang, Jingge and Feng, Tao and Ma, Fei and Li, Yang},
  booktitle={2025 IEEE 22nd International Symposium on Biomedical Imaging (ISBI)},
  pages={1--5},
  year={2025},
  organization={IEEE}
}

@inproceedings{shou2025dynamic,
  title={Dynamic graph neural ode network for multi-modal emotion recognition in conversation},
  author={Shou, Yuntao and Meng, Tao and Ai, Wei and Li, Keqin},
  booktitle={Proceedings of the 31st International Conference on Computational Linguistics},
  pages={256--268},
  year={2025}
}

@article{shou2025multimodal,
  title={Multimodal large language models meet multimodal emotion recognition and reasoning: A survey},
  author={Shou, Yuntao and Meng, Tao and Ai, Wei and Li, Keqin},
  journal={arXiv preprint arXiv:2509.24322},
  year={2025}
}

@inproceedings{liu2024improved,
  title={Improved baselines with visual instruction tuning},
  author={Liu, Haotian and Li, Chunyuan and Li, Yuheng and Lee, Yong Jae},
  booktitle={Proceedings of the IEEE/CVF Conference on Computer Vision and Pattern Recognition},
  pages={26296--26306},
  year={2024}
}

@inproceedings{yao2024advancing,
  title={Advancing semi-supervised EEG emotion recognition through feature extraction with mixup and large language models},
  author={Yao, Shiyi and Liu, Longfei and Lu, Jingyi and Wu, Dan and Li, Ye},
  booktitle={2024 ieee international conference on bioinformatics and biomedicine (bibm)},
  pages={2772--2779},
  year={2024},
  organization={IEEE}
}

@article{deng2025mind2matter,
  title={Mind2matter: Creating 3d models from eeg signals},
  author={Deng, Xia and Chen, Shen and Zhou, Jiale and Li, Lei},
  journal={arXiv preprint arXiv:2504.11936},
  year={2025}
}

@inproceedings{liu2024hidden,
  title={Hidden States in LLMs Improve EEG Representation Learning and Visual Decoding.},
  author={Liu, Aoyang and Jing, Haodong and Liu, Yulong and Ma, Yongqiang and Zheng, Nanning},
  booktitle={ECAI},
  volume={392},
  pages={2130--2137},
  year={2024}
}

@inproceedings{mishra2025thought2text,
  title={Thought2Text: text generation from EEG signal using large language models (LLMs)},
  author={Mishra, Abhijit and Shukla, Shreya and Torres, Jose and Gwizdka, Jacek and Roychowdhury, Shounak},
  booktitle={Findings of the Association for Computational Linguistics: NAACL 2025},
  pages={3747--3759},
  year={2025}
}

@article{feng2025stftcodec,
  title={STFTCodec: High-Fidelity Audio Compression through Time-Frequency Domain Representation},
  author={Feng, Tao and Zhao, Zhiyuan and Xie, Yifan and Ye, Yuqi and Luo, Xiangyang and Guan, Xun and Li, Yu},
  journal={arXiv preprint arXiv:2503.16989},
  year={2025}
}

@article{yang2023gpt4tools,
  title={Gpt4tools: Teaching large language model to use tools via self-instruction},
  author={Yang, Rui and Song, Lin and Li, Yanwei and Zhao, Sijie and Ge, Yixiao and Li, Xiu and Shan, Ying},
  journal={Advances in Neural Information Processing Systems},
  volume={36},
  pages={71995--72007},
  year={2023}
}

@inproceedings{jianglarge,
  title={Large Brain Model for Learning Generic Representations with Tremendous EEG Data in BCI},
  author={Jiang, Weibang and Zhao, Liming and Lu, Bao-liang},
  booktitle={The Twelfth International Conference on Learning Representations},
  year={2024}
}

@article{wang2024eegpt,
  title={Eegpt: Pretrained transformer for universal and reliable representation of eeg signals},
  author={Wang, Guangyu and Liu, Wenchao and He, Yuhong and Xu, Cong and Ma, Lin and Li, Haifeng},
  journal={Advances in Neural Information Processing Systems},
  volume={37},
  pages={39249--39280},
  year={2024}
}

@article{xie2025universal,
  title={Universal Visuo-Tactile Video Understanding for Embodied Interaction},
  author={Xie, Yifan and Li, Mingyang and Li, Shoujie and Li, Xingting and Chen, Guangyu and Ma, Fei and Yu, Fei Richard and Ding, Wenbo},
  journal={arXiv preprint arXiv:2505.22566},
  year={2025}
}

@article{lian2024affectgpt,
  title={AffectGPT: Dataset and framework for explainable multimodal emotion recognition},
  author={Lian, Zheng and Sun, Haiyang and Sun, Licai and Yi, Jiangyan and Liu, Bin and Tao, Jianhua},
  journal={arXiv preprint arXiv:2407.07653},
  year={2024}
}

@article{yang2025qwen3,
  title={Qwen3 technical report},
  author={Yang, An and Li, Anfeng and Yang, Baosong and Zhang, Beichen and Hui, Binyuan and Zheng, Bo and Yu, Bowen and Gao, Chang and Huang, Chengen and Lv, Chenxu and others},
  journal={arXiv preprint arXiv:2505.09388},
  year={2025}
}

@article{loshchilov2017decoupled,
  title={Decoupled weight decay regularization},
  author={Loshchilov, Ilya and Hutter, Frank},
  journal={arXiv preprint arXiv:1711.05101},
  year={2017}
}

@article{jiang2024neurolm,
  title={NeuroLM: A universal multi-task foundation model for bridging the gap between language and EEG signals},
  author={Jiang, Wei-Bang and Wang, Yansen and Lu, Bao-Liang and Li, Dongsheng},
  journal={arXiv preprint arXiv:2409.00101},
  year={2024}
}

@article{huang2023discrepancy,
  title={Discrepancy between inter-and intra-subject variability in EEG-based motor imagery brain-computer interface: Evidence from multiple perspectives},
  author={Huang, Gan and Zhao, Zhiheng and Zhang, Shaorong and Hu, Zhenxing and Fan, Jiaming and Fu, Meisong and Chen, Jiale and Xiao, Yaqiong and Wang, Jun and Dan, Guo},
  journal={Frontiers in neuroscience},
  volume={17},
  pages={1122661},
  year={2023},
  publisher={Frontiers Media SA}
}

@article{babu2025large,
  title={Large Language Models for EEG: A Comprehensive Survey and Taxonomy},
  author={Babu, Naseem and Mathew, Jimson and Vinod, AP},
  journal={arXiv preprint arXiv:2506.06353},
  year={2025}
}

@article{chen2025eeg,
  title={EEG emotion copilot: Optimizing lightweight LLMS for emotional EEG interpretation with assisted medical record generation},
  author={Chen, Hongyu and Zeng, Weiming and Chen, Chengcheng and Cai, Luhui and Wang, Fei and Shi, Yuhu and Wang, Lei and Zhang, Wei and Li, Yueyang and Yan, Hongjie and others},
  journal={Neural Networks},
  pages={107848},
  year={2025},
  publisher={Elsevier}
}

@article{shen2023hugginggpt,
  title={Hugginggpt: Solving ai tasks with chatgpt and its friends in hugging face},
  author={Shen, Yongliang and Song, Kaitao and Tan, Xu and Li, Dongsheng and Lu, Weiming and Zhuang, Yueting},
  journal={Advances in Neural Information Processing Systems},
  volume={36},
  pages={38154--38180},
  year={2023}
}

@article{kingma2014adam,
  title={Adam: A method for stochastic optimization},
  author={Kingma, Diederik P},
  journal={arXiv preprint arXiv:1412.6980},
  year={2014}
}

@inproceedings{xie2025audio,
  title={Audio-Driven Talking Face Video Generation with Joint Uncertainty Learning},
  author={Xie, Yifan and Ma, Fei and Bin, Yi and He, Ying and Yu, Fei},
  booktitle={Proceedings of the 2025 International Conference on Multimedia Retrieval},
  pages={1588--1597},
  year={2025}
}

@article{ma2025generative,
  title={Generative technology for human emotion recognition: A scoping review},
  author={Ma, Fei and Yuan, Yucheng and Xie, Yifan and Ren, Hongwei and Liu, Ivan and He, Ying and Ren, Fuji and Yu, Fei Richard and Ni, Shiguang},
  journal={Information Fusion},
  volume={115},
  pages={102753},
  year={2025},
  publisher={Elsevier}
}

@inproceedings{xie2025pointtalk,
  title={Pointtalk: Audio-driven dynamic lip point cloud for 3d gaussian-based talking head synthesis},
  author={Xie, Yifan and Feng, Tao and Zhang, Xin and Luo, Xiangyang and Guo, Zixuan and Yu, Weijiang and Chang, Heng and Ma, Fei and Yu, Fei Richard},
  booktitle={Proceedings of the AAAI Conference on Artificial Intelligence},
  volume={39},
  number={8},
  pages={8753--8761},
  year={2025}
}




\end{document}